%% file: main.tex
\documentclass[A4paper, conference]{IEEEtran} 
\IEEEoverridecommandlockouts
\usepackage{graphicx}
\usepackage{amsmath,amsthm,amsfonts,amssymb}
\usepackage{cite}
\usepackage{bm}
\usepackage{bbm}
\usepackage{url}
\usepackage{array}
\usepackage{float}
\usepackage{color}
\usepackage{multirow}
\usepackage{booktabs}
\usepackage{xcolor}
\usepackage{enumerate}
\usepackage{epstopdf}
\usepackage{placeins}
\usepackage{tabularx}
\usepackage{algorithm}
\usepackage[english]{babel}
\usepackage{algpseudocode}
\usepackage{listings}
\usepackage{tcolorbox}
\usepackage{pdfpages}
\tcbset{sharp corners, boxrule=0.5pt, top=0mm, bottom=0pt, left=0pt, right=0pt, boxsep=0.5pt}
\usepackage{hyperref}
\usepackage{lstautogobble}
\usepackage[caption=false,font=scriptsize,labelfont=rm,textfont=rm]{subfig}
\tcbuselibrary{skins, breakable}

\definecolor{vgreen}{RGB}{104,180,104}
\definecolor{vblue}{RGB}{49,49,255}
\definecolor{vorange}{RGB}{255,143,102}

\makeatletter
\newcommand*\@lbracket{[}
\newcommand*\@rbracket{]}
\newcommand*\@colon{:}
\newcommand*\colorIndex{%
    \edef\@temp{\the\lst@token}%
    \ifx\@temp\@lbracket \color{black}%
    \else\ifx\@temp\@rbracket \color{black}%
    \else\ifx\@temp\@colon \color{black}%
    \else \color{vorange}%
    \fi\fi\fi
}
\makeatother

\usepackage{trace}

\begin{document}
\title{\makebox[\linewidth]{\parbox{\dimexpr\textwidth+0.00cm\relax}{\centering Rephrase and Contrast: Fine-Tuning Language Models for Enhanced Understanding of Communication and Computer Networks}}}

\author{
    \IEEEauthorblockN{Liujianfu Wang$^{a\dagger}$, Yuyang Du$^{a\dagger}$, Jingqi Lin$^{b\dagger}$, Kexin Chen$^a$, Soung Chang Liew$^{a*}$}
    \IEEEauthorblockA{$^a$ The Chinese University of Hong Kong, Hong Kong SAR, China}
    \IEEEauthorblockA{$^b$ Huazhong University of Science and Technology, Wuhan, China}
    %\IEEEauthorblockA{$^c$ Department of Computer Science Engineering,  Chinese University of Hong Kong, Hong Kong SAR, China}
    \vspace{-5em}
    \IEEEauthorblockA{\thanks{This work was partially supported by the Shen Zhen-Hong Kong-Macao technical program (Type C) under Grant No. SGDX20230821094359004.}}
    \IEEEauthorblockA{\thanks{Code, data, and model availability: https://github.com/1155157110/RaC.}}
    \IEEEauthorblockA{\thanks{$^\dagger$L. Wang, Y. Du, J. Lin contribute equally. This work was completed during J. Lin's internship at the Chinese University of Hong Kong.}}
    \IEEEauthorblockA{\thanks{$^*$S. C. Liew (soung@ie.cuhk.edu.hk) is the corresponding author.}}
}
\maketitle

\begin{abstract}
Large language models (LLMs) are being widely researched across various disciplines, with significant recent efforts focusing on adapting LLMs for understanding of how communication networks operate. However, over-reliance on prompting techniques hinders the full exploitation of the generalization ability of these models, and the lack of efficient fine-tuning methods prevents the full realization of lightweight LLMs’ potential. This paper addresses these challenges by introducing our Rephrase and Contrast (RaC) framework, an efficient fine-tuning framework. RaC enhances LLMs' comprehension and critical thinking abilities by incorporating question reformulation and contrastive analysis of correct and incorrect answers during the fine-tuning process. Experimental results demonstrate a 63.73\% accuracy improvement over the foundational model when tested on a comprehensive networking problem set. Moreover, to efficiently construct the dataset for RaC fine-tuning, we develop a GPT-assisted data mining method for generating high-quality question-answer (QA) pairs; furthermore, we introduce ChoiceBoost, a data augmentation technique that expands dataset size while reducing answer-order bias. Apart from these technical innovations, we contribute to the networking community by open-sourcing valuable research resources, including: 1) the fine-tuned networking model referred to as RaC-Net, 2) the training dataset used for fine-tuning the model, 3) three testing problem sets of different difficulties to serve as benchmarks for future research, and 4) code associated with the above resources. 
\end{abstract}

\begin{IEEEkeywords}
Large language model, wireless/wired networks supervised fine-tuning, benchmark dataset
\end{IEEEkeywords}

\section{Introduction}\label{sec-I}
\input{./Sources/Section_I.tex}

\section{Related Work}\label{sec-II}
\input{./Sources/Section_II.tex}

\section{RaC: An Efficient Fine-tuning Framework}\label{sec-III}
\input{./Sources/Section_III.tex}

\section{Training and Testing Datasets}\label{sec-IV}
\input{./Sources/Section_IV.tex}

\section{Experiments}\label{sec-V}
\input{./Sources/Section_V.tex}

\section{Conclusion}\label{sec-VI}
\input{./Sources/Section_VI.tex}

\section{Acknowledgment}
This work was supported in part by the Shen Zhen-Hong Kong-Macao technical program under Grant No. SGDX20230821094359004.

\bibliographystyle{IEEEtran}
\bibliography{main}

 \newpage
 \section*{Appendix A}
 \nopagebreak
 \includegraphics[width=\textwidth]{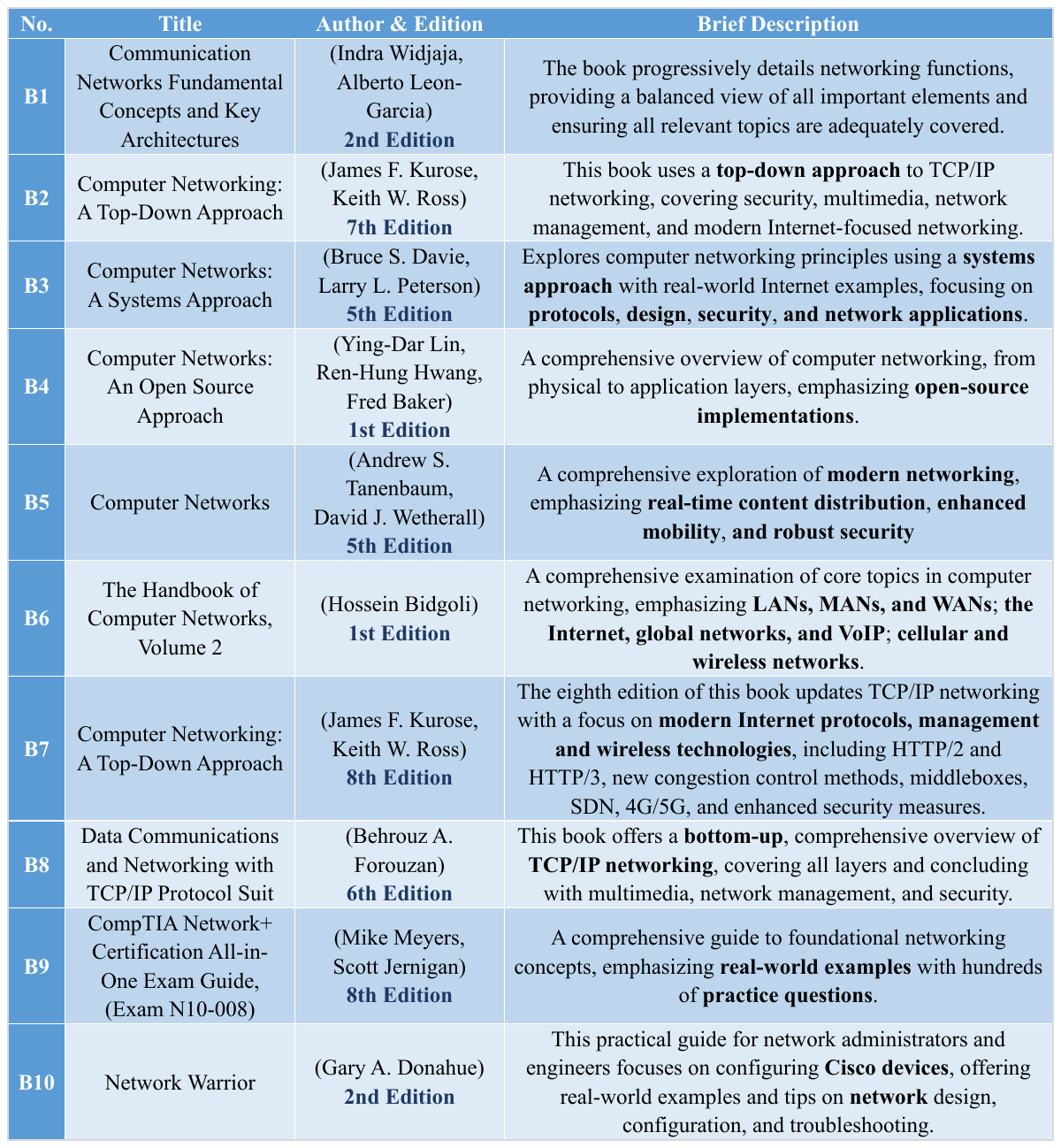}

\end{document}

%% file: Sources/Section_I.tex
In recent years, artificial intelligence has witnessed a paradigm shift with the advent of large language models (LLMs) that exhibit near-human intelligence across various domains \cite{niu2024large}. These models have captured the attention of researchers and practitioners from diverse fields. The remarkable capabilities of LLMs have sparked a surge of interest in their potential applications across various scientific disciplines.

The field of communication networks is no exception to this trend. Recent studies have explored the application of LLMs in several key areas within the networking domain. For instance, researchers have investigated the use of LLMs in network topology analysis \cite{he2024llm, mani2023enhancing}, wireless device configuration \cite{mondal2023llms, lian2023configuration}, network setup optimization \cite{cui2023llmind, wu2024netllm, shao2024wirelessllm}, network diagnosis \cite{kotaru2023adapting, zhou2023towards}, and network security \cite{meng2024large}. These efforts highlight the potential of LLMs to enhance the performance and efficiency of various network-related tasks.

Despite the considerable efforts, current approaches to integrating LLMs in communication networks suffer from several drawbacks. One drawback is the use of prompt tuning techniques. While prompt tuning methods such as chain of thought (CoT) enable rapid proof-of-concept demonstrations, they are inherently inefficient and often clumsy. Crafting effective prompts may require extensive trial-and-error efforts, and the resulting solutions may lack robustness and generalizability because the foundational parameters within LLMs are not modified to tailor for networking tasks. Moreover, the computational overhead associated with prompt tuning can be substantial, especially for large-scale network applications \cite{zou2024promptintern}.

Fine-tuning methods that tailor the parameters of LLMs for networking tasks have the potential to overcome such drawbacks of prompt engineering. This paper introduces an efficient LLM fine-tuning scheme referred to as the Rephrase and Contrast (RaC) framework. Unlike conventional fine-tuning methods that only use question-answer (QA) pairs, RaC supplies the LLM with additional information, including 1) a reformulation of the question asked, and 2) a contrastive analysis of both the correct answer and potential incorrect answers. This approach, which effectively mimics how humans learn, enhances problem comprehension by LLMs and significantly improves their critical thinking ability essential for analyzing complex networking issues. Experimental results on RaC indicate a 63.73\% accuracy improvement compared to the foundational model without fine-tuning when tested on a comprehensive networking problem set.

Apart from the RaC fine-tuning framework, we notice a lack of datasets for training LLMs on networking tasks. To address this, we develop a GPT-assisted data mining method and introduce ChoiceBoost, a simple yet highly effective data augmentation technique for building training datasets.

For data mining, we leverage the language processing ability of GPT-4 to analyze raw materials extracted from classic networking textbooks and generate high-quality QA pairs that meet the needs of RaC fine-tuning. The QA pair generation process applies the in-context learning (ICL) technique to guide the GPT-4 model in structuring the data and help the model better understand what output is desired in each step. 

For data augmentation, ChoiceBoost augments the dataset by altering the order of the correct answer and incorrect answers in multiple-choice questions. This approach expands the dataset size by four times, maintaining the same knowledge content while offering varied choice arrangements. ChoiceBoost effectively reduces bias related to answer order and reinforces the model's understanding of the material. Subsequent ablation studies show a clear improvement in the model's performance attributed to ChoiceBoost.

Beyond our technical contributions (i.e., RaC, the GPT-assisted data mining scheme, and ChoiceBoost), we are also releasing several valuable resources to the networking community. These include: 1) our training dataset used for LLM fine-tuning, 2) three novel testing problem sets that can serve as testing benchmarks for future research, 3) the fine-tuned models referred to as the RaC-Net, and 4) all code associated with the above three components. This comprehensive release will enable researchers and practitioners to reproduce our results and build upon our work to develop advanced network-oriented LLMs.

%% file: Sources/Section_II.tex
Recent advances in LLMs have inspired many researchers to explore their adaptation for networking tasks \cite{he2024llm, mani2023enhancing, mondal2023llms, lian2023configuration, cui2023llmind, wu2024netllm, shao2024wirelessllm, kotaru2023adapting, zhou2023towards, meng2024large, nikbakht2024tspec, maatouk2023teleqna}. This section highlights this paper’s unique contributions compared with these previous works. Our comparison centers on two key aspects: 1) the methodologies employed, and 2) the availability of components associated with projects, including codes, training data, and testing problem sets.
 
\textbf{Methodologies}: Most previous works were built upon prompt tuning over off-the-shelf models with only a few considered model fine-tuning. For example, in \cite{shao2024wirelessllm}, the authors put forth a framework that empowers LLMs with various model-optimizing techniques including model fine-tuning. However, the work only conducted experiments on prompt-tuning methods. In \cite{he2024llm, mondal2023llms, lian2023configuration, kotaru2023adapting, zhou2023towards, meng2024large, shao2024wirelessllm, cui2023llmind}, few-shot prompting methods, such as in-context learning (ICL), are used to guide LLMs to generate answers reliably. Chain of thought (CoT) was applied in \cite{he2024llm, zhou2023towards, cui2023llmind, shao2024wirelessllm} to enhance LLMs’ reasoning ability. Additionally, retrieval methods like retrieval augmented generation (RAG) are used in \cite{nikbakht2024tspec, shao2024wirelessllm} to enhance the off-the-shelf LLM’s knowledge in wireless networks.

These off-the-shelf models were not specifically trained for networking applications. Their expertise is in conventional NLP tasks, such as language translation, rather than in networking. Despite the advanced prompting techniques to enhance LLMs' comprehension of networks, their efficacy is limited due to the inherent limitations in LLMs’ foundational architectures.

Within the small handful of works that adapted LLMs through fine-tuning for wireless networking tasks, \cite{wu2024netllm} adapted a LLaMA-2 7B model with the low-rank adaptation (LoRA) technique and demonstrated the model’s performance in specific tasks such as adaptive bit rate streaming and cluster job scheduling. However, \cite{wu2024netllm} did not make new contributions in fine-tuning methods: it directly used the LoRA method without further modification to enhance its performance.
 
Unlike the above previous work, our paper put forth the RaC scheme for effective LLM fine-tuning (see Section III for details). Furthermore, to complement, we develop a GPT-assisted data mining method and a data augmentation scheme to gloom the dataset used to fine-tune LLMs (see Section IV-A for details). These methods make possible a more efficient utilization of the knowledge within the training dataset.

\textbf{Project Components}: In previous work, only \cite{mani2023enhancing}, \cite{lian2023configuration}, \cite{nikbakht2024tspec}, and \cite{maatouk2023teleqna} released their research data. However, the released datasets are either too small for LLM adaptations \cite{mani2023enhancing, lian2023configuration}, or cannot be directly used in model fine-tuning, as the corresponding code for data post-processing and scripts for fine-tuning models are not released \cite{nikbakht2024tspec, maatouk2023teleqna}. Compared with previous research, we provide a dataset that is self-contained as far as LLM adaptation is concerned. Additionally, we also provide the community with the source code associated with the model fine-tuning, facilitating reproductions and follow-ups of this work.

In addition, recognizing the absence of a generally accepted testing benchmark for evaluating the performance of LLMs on networking tasks, we release three testing problem sets of different difficulties to serve as network-oriented LLM evaluation baselines.

%% file: Sources/Section_III.tex
This section introduces RaC, a framework designed to enhance the efficiency of model fine-tuning. RaC is inspired by two observations:

\textbf{Observation 1}: The Significance of Rephrasing. In human society, question rephrasing plays a crucial role in ensuring accurate and efficient communication. When faced with an unfamiliar question, we often rephrase it to eliminate ambiguities. Additionally, providing the detailed context of the question deepens the understanding by the responder to understand.

\textbf{Observation 2}: Wrong Answer Matters. In human learning, analyzing incorrect answers enhances understanding. By contrasting them with correct answers, we gain a deeper understanding of the topic, fostering critical thinking. Additionally, explanations of wrong answers can introduce new information not covered in the correct answer's explanation.

\begin{figure*}[htbp]
  \centering
  \includegraphics[width=\textwidth]{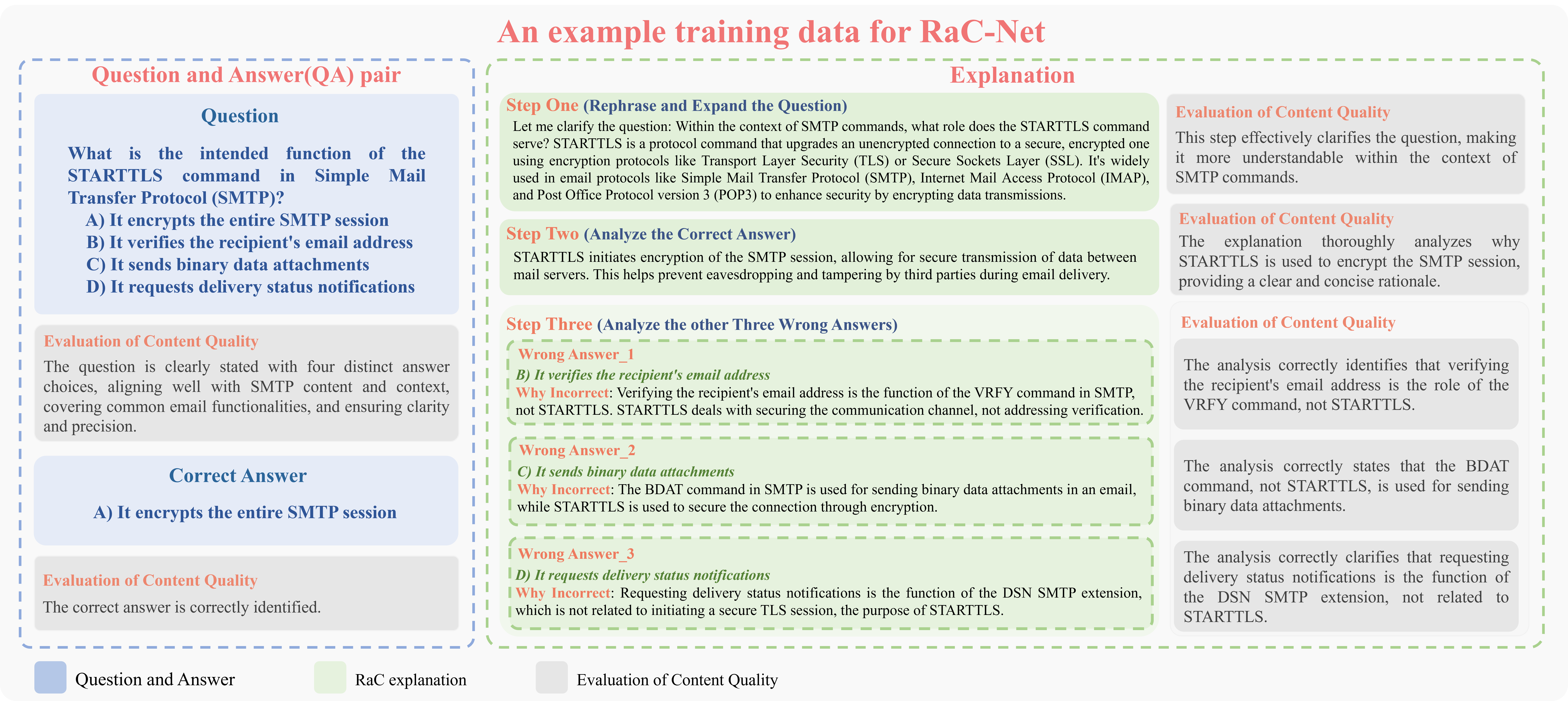}\\
  \captionsetup{font={footnotesize}}
  \caption{An example of training data that meets the data structure requirements of our RaC framework. For easier understanding and better assessment of data quality, comments have been added in the grey box. These comments are not part of the training dataset; they are included for illustrative purposes only.}\label{fig1}
\end{figure*}

Based on these observations, our RaC framework supplements the data used to fine-tune LLMs by rephrasing questions and performing contrastive analysis of correct and incorrect answers. Fig. \ref{fig1} provides an example to illustrate the process.

Before delving into Fig. \ref{fig1}, let us briefly explain why we use the multiple-choice format as the target QA format in RaC-Net. This format offers an objective and easy-to-grade method for evaluating the model's accuracy. Additionally, it allows for analysis of the model's decision-making process by assessing its ability to identify the best answer among multiple plausible options.

In Fig. \ref{fig1}, the multiple-choice question asks about the intended function of the STARTTLS command, commonly used in the Simple Mail Transfer Protocol (SMTP). The correct answer is A) It encrypts the entire SMTP session.

Using the RaC method, we first rephrase the question by clarifying the term "intended function" and pointing out that STARTTLS is a command that upgrades an existing, unencrypted connection to a secure, encrypted one using TLS or SSL. It is widely used in email protocols SMTP, IMAP, and POP3 to enhance security by encrypting data transmissions.

We then move on to contrastive analysis of the correct and incorrect answers. We explain that the correct answer identifies STARTTLS as the command that encrypts the SMTP session, emphasizing its role in securing email transmissions by preventing third-party eavesdropping and tampering. In contrast, each incorrect answer mistakenly assigns the functions of other SMTP commands to STARTTLS, such as VRFY for verifying email addresses, BDAT for sending binary data, and DSN for requesting delivery status. This analysis effectively differentiates between various SMTP commands, highlighting STARTTLS's specific security role. It is worth noting that all of the question reformulation contents and contrastive analysis are autonomously generated by the GPT model, without human intervention.

For the RaC implementation, our framework can be adopted in a variety of supervised fine-tuning methods. We implement LoRA fine-tune to verify the performance of our RaC framework. Compared with LoRA without RaC, our RaC framework improves models' comprehension in the fine-tuning process.

%% file: Sources/Section_IV.tex
The extensive dataset required for LLM fine-tuning necessitates an automated approach for QA pair generation. We leverage the GPT-4 model to generate a substantial corpus of high-quality QA pairs. Importantly, we highlight that the QA pair we are looking for is more than a simple combination of a question and an answer. A core feature of our QA pair is the inclusion of problem rephrasing and contrasting statements, which are required in the RaC framework. The rest of this paper refers to the new QA pair as the RaC QA pair to distinguish it from the conventional QA pair.

In the following, Subsection A presents technical details about the data generation process, including our GPT-assisted data mining scheme and ChoiceBoost, the rotation-based data augmentation method designed for the multiple-choice setup in RaC. Subsection B introduces our open-sourced training dataset and testing problem sets for future research endeavors.

\subsection{Dataset Construction}
Our dataset is built upon 10 textbooks on computer networking, which include both foundational theories and the latest technical advancements in the networking field. Each textbook was chosen for its unique technical insights, thus ensuring comprehensive coverage of the networking field. A detailed description of these books is provided in Appendix A.

In our dataset, each instance is formatted as a multiple-choice QA pair, including one correct answer, three incorrect answers, problem rephrase, and explanations associated with these four choices. This data structure aligns well with the RaC fine-tuning framework. The following procedure was applied to ensure an efficient and high-quality data mining process.

\textbf{Preprocessing}: We employed optical character recognition (OCR) software to convert the selected textbooks into textual material. Along with the OCR process, we removed elements incompatible with the GPT-4 model, such as images and their captions. We segmented the textual content by section to accommodate GPT-4's token limit without harming the context coherence within each segment. After purification and segmentation, the resulting textual context of each book section was structured into the JSON format for subsequent processing.

\begin{figure}[htbp]
  \centering
  \includegraphics[width=0.46\textwidth]{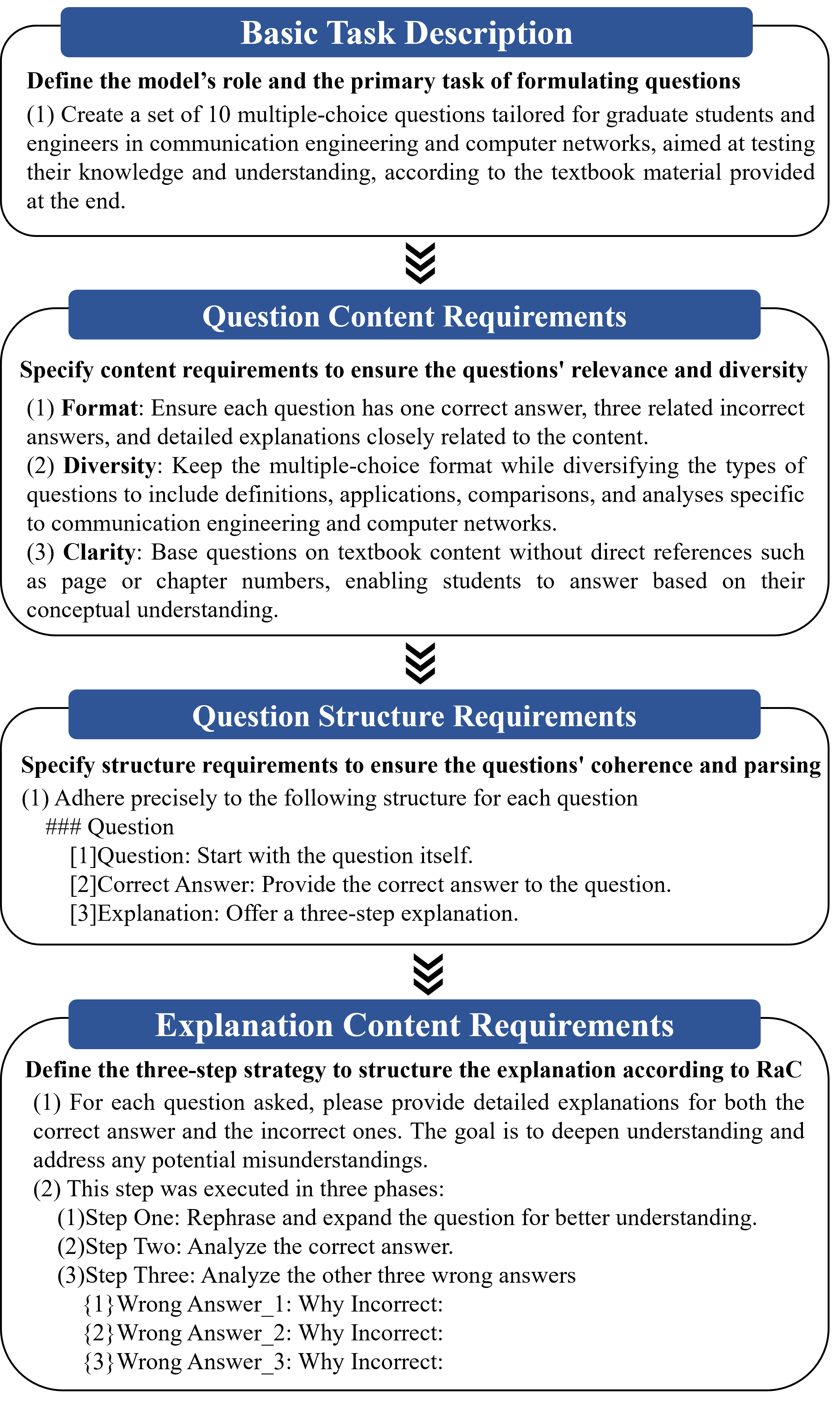}\\
  \captionsetup{font={footnotesize}}
  \caption{LLM prompt template used by RaC-Net for creating QA pairs.}\label{fig2}
\end{figure}

\textbf{RaC QA Pair Generation}: We leveraged the GPT-4 API for the implementation of QA pair generation, with the following configuration parameters used: temperature = 1.0, top-p = 1.0, frequency penalty = 0.0, and presence penalty = 0.0. Fig. \ref{fig2} illustrates the detailed prompt employed in this QA pair generation process. The process begins by describing the basic task for the model, which specifies the creation of questions targeting networking knowledge. Then the prompt outlines the requirements for question content and structure. Finally, it defines a three-step strategy for generating the RaC explanations.
 
\textbf{Manual Review}: Following QA pair generation, we randomly sampled 200 QA pairs for evaluation. Each selected QA pair underwent a manual review. If any component (including question, answer, or explanation) was found to be inaccurate or illogical, we returned to the prompt design and revised it accordingly. This iterative process continued until the dataset met our stringent quality criteria, i.e., meticulously crafted questions, exact answers, and logically coherent, structurally sound explanations.

After the GPT-assisted QA pair generation, we now explain how we augment the obtained dataset with ChoiceBoost. Analysis of the generated data reveals two significant observations: 1) While GPT-4’s assistance facilitates the production of high-quality data, the associated API costs would be prohibitive for large-scale data expansion. 2) The generated data exhibited potential bias in option selection, which could mislead the fine-tuned model into acquiring unfounded statistical bias.  

ChoiceBoost is designed to mitigate the above two issues. In ChoiceBoost, each question in the dataset was duplicated four times, with each duplicate assigned a different correct answer (A, B, C, or D), while maintaining the original order of incorrect options. This approach substantially expands the dataset without additional API cost (which enhances the model's capacity by learning from diverse answer configurations) and mitigates biases inherent in the initial data generation process.

\subsection{Data Resources Released}
The initial dataset derived from ten textbooks without post-processing data augmentation comprises 15,119 QA pairs. These original QA pairs encompass essential sub-domains knowledge of computer networking described in each individual textbook, such as layered network architecture, protocols design, network security, and network management. From the raw data, 756 QA pairs (approximately 5\%) were randomly selected to form a testing problem set. We then used ChoiceBoost to augment the remaining 15,119 – 756 = 14,363 QA pairs, resulting in an unbiased training dataset containing 14,363 * 4 = 57,452 QA pairs. 

In addition to the GPT-generated QA pairs, we manually collected 327 multiple-choice testing questions closely aligned with the content of the ten selected textbooks. These questions were sourced from publicly available resources, including open-source exam papers in online databases and problem sets provided within the textbooks. Note that these QA pairs are exclusively used for testing purposes only, and they lack problem rephrasing and answer explanations required in the RaC framework. The value of the new testing problem set lies in its focus on complex logical reasoning and intricate calculations. For instance, such problems might require calculating the maximum network throughput under changing bandwidth conditions, or involve multi-step reasoning, such as optimizing a routing algorithm considering traffic fluctuations and delay constraints. The motivation for building the more challenging dataset stems from our observation that GPT-generated QA pairs may not adequately address complex reasoning and mathematical computations due to the limitations of the GPT-4 model in these areas. Instead, the GPT-generated pairs primarily emphasize concept comprehension. For example, a typical problem from the set might ask, 'What is the function of a router in a network?' or 'Which OSI layer is responsible for error detection and correction?'. While the 756 GPT-generated testing QA pairs can assess a model's grasp of networking concepts, a testing problem set concentrating on logical reasoning and mathematical computation is equally essential for a comprehensive evaluation of the fine-tuned model. For the rest of this paper, we refer to the testing problem set containing 756 GPT-generated problems as the "easy problem set" and the 327 manually collected problems as the "hard problem set."

\begin{figure*}[htbp]
  \centering
  \includegraphics[width=\textwidth]{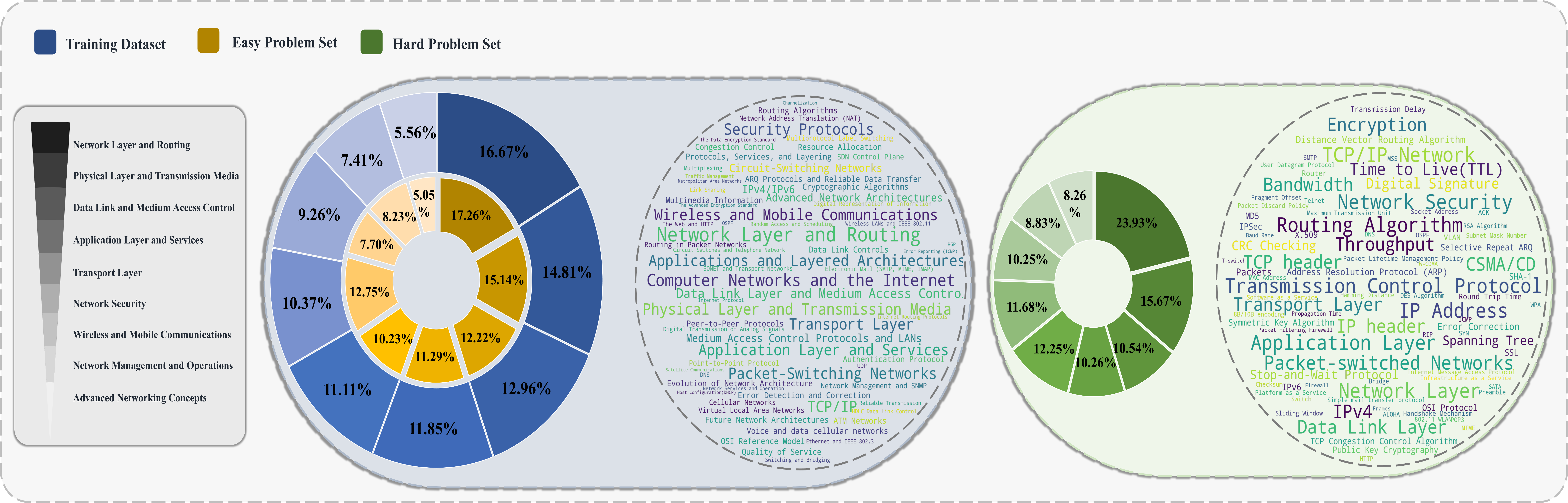}\\
  \captionsetup{font={footnotesize}}
  \caption{An overview of the released data resource. The three pie charts share the same components listed in the left legend. The legend uses a grayscale graph to indicate each component, with the darker shade corresponding to the deeper-color segments in pie charts (e.g., the darkest blue, orange, and green sections match the darkest gray category, which represents the Network layer and Routing).}\label{fig3}
\end{figure*}

Fig. \ref{fig3} visually illustrates the data distribution of the training dataset and the easy/hard problem set. In this figure, networking knowledge is categorized into nine distinct sub-domains. The figure shows that, the training data, the easy test problem sets, and the hard test problems, all exhibit a balanced coverage of these sub-domains, thereby ensuring that the model is trained and evaluated with a comprehensive and generalized focus on the networking field. These two word clouds further detail the sub-domain content distribution. The left word cloud highlights the foundational and all-inclusive topics in the training dataset and easy problem set; the right one, on the other hand, reflects the hard problem set focuses on more complex and detailed aspects of networking.

The easy and hard problem sets constitute two critical testing benchmarks that assess an LLM's performance in the networking domain from contrasting perspectives. An additional crucial aspect of the model’s evaluation is its performance in a comprehensive and practical problem set with both easy and hard problems. To this end, we randomly down-sampled the easy problem set to create a subset comparable in size to the hard dataset (i.e., 327 QA pairs). Subsequently, we merged this new subset with the hard problem set to form a comprehensive dataset. All three problem sets (easy, hard, and comprehensive) are used in Section V for model evaluation.

%% file: Sources/Section_V.tex
\begin{figure}[htbp]
  \centering
  \includegraphics[width=0.46\textwidth]{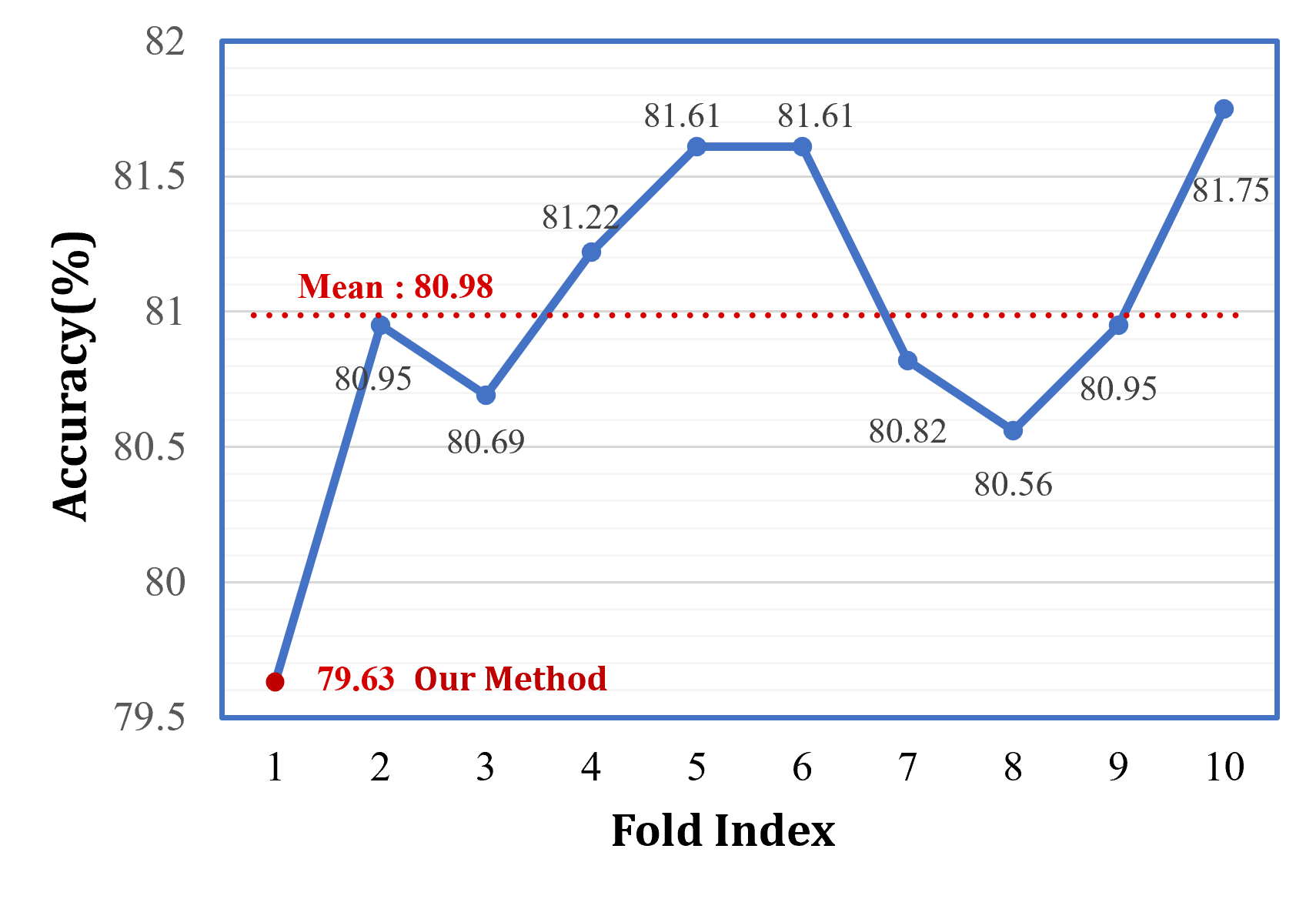}\\
  \captionsetup{font={footnotesize}}
  \caption{Accuracies of models of fold index 1 to 10 tested on the easy problem set. Due to the limitation in our computational resources, we randomly selected 10 folds among the 20 candidate folds for concept proofing.}\label{fig4}
\end{figure}

To assess the robustness and generalizability of our proposed method, we conducted a k-fold cross-validation experiment. K-fold cross-validation is a statistical technique used to evaluate machine learning models by partitioning the data into k subsets, using k-1 subsets for training and the remaining subset for testing, and then repeating this process k times with different test subsets. In section \ref{sec-IV} we detail the constitution of training data and test data. As we split the GPT-generated QA pair into training data and testing problem set with a ratio of 19:1, we opted for a 20-fold cross-validation approach (k = 20) to ensure comprehensive evaluations, where folds do not overlap with one another. In each iteration, one fold was designated as the test set, while the remaining 19 folds constituted the training data, to which data augmentation techniques were applied before model training. With respect to the explanation given in Section IV.B, this methodology results in approximately 756 data for the test set and 57,452 data (14,363 * 4) for the training set in each k-fold iteration. This approach allows us to utilize all available data for both training and testing, thus providing a more reliable estimate of the model's performance. For each iteration, data augmentation techniques were applied exclusively to the training set, while the corresponding test set remained unaltered to maintain the integrity of the evaluation process. Fig. \ref{fig4} illustrates the accuracy metrics obtained for each fold.

For the experimental setup, we employed the LLaMA-2-7B model as our foundational language model and applied LoRA fine-tuning. The training process was conducted over 35 epochs with a learning rate of 1e-4. Computational resources consisted of four NVIDIA A6000 GPU cards to facilitate efficient parallel processing.

\begin{figure*}[htbp]
  \centering
  \includegraphics[width=0.85\textwidth]{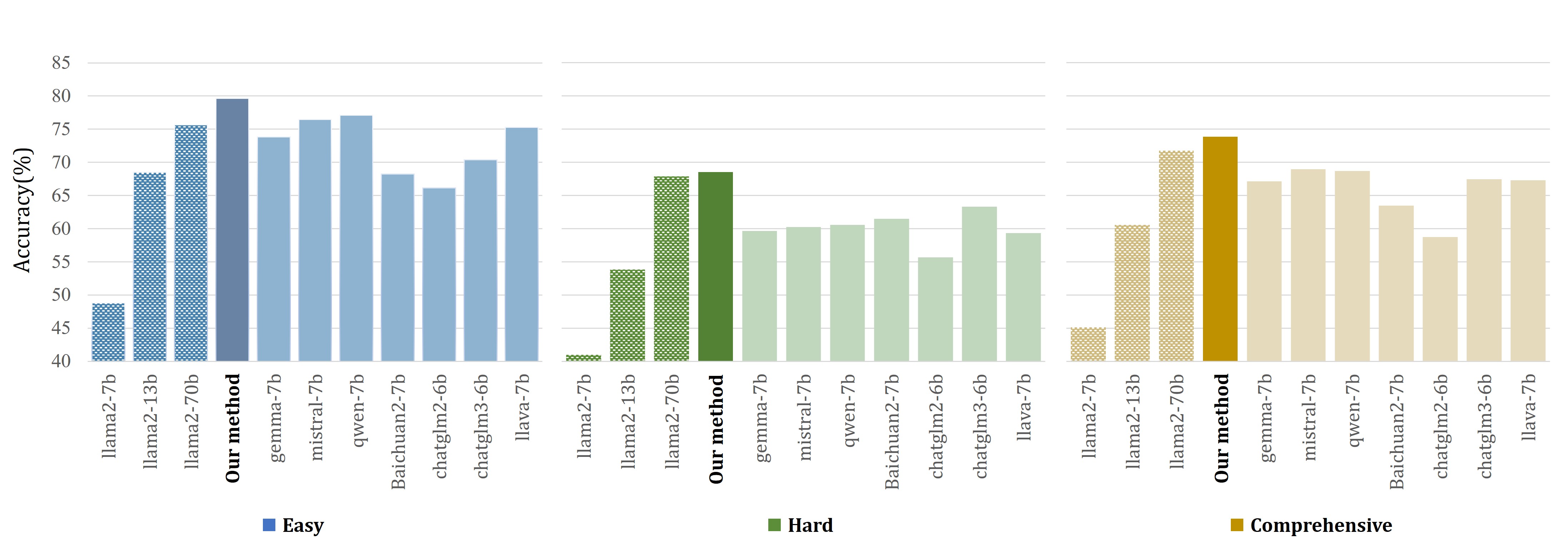}\\
  \captionsetup{font={footnotesize}}
  \caption{Accuracies of baseline models tested on easy, hard, and comprehensive testing problem sets.}\label{fig5}
\end{figure*}

The results demonstrate a consistent performance across all folds, with a mean accuracy of 80.98\% accuracy across different folds and minimal variance in accuracy scores. This uniformity suggests that our method's efficacy is not contingent upon specific dataset configurations but rather represents a generalizable solution. The observed consistency across diverse data subsets underscores the robustness and broad applicability of our proposed approach. Given the consistent performance between each fold, we here select the trained model of fold \#1 as our method to compare with other models.

We benchmark our method with different baselines to demonstrate the superiority of our model in the networking domain. Specifically, we compare our method with LLaMA-2 models of various sizes (i.e., 7B, 13B, and 70B) and representative off-the-shelf LLMs that are 1) released at a similar time as our LLaMA-2 7B foundational model does, and 2) of similar parameter sizes. Fig. \ref{fig5} presents the accuracy of our model and all the above baseline models on the three testing problem sets we built in Section IV-B.

For evaluation results on the easy testing problem set, our model correctly answers 80\% of questions, significantly surpassing all other baseline models in terms of accuracy. The original LLaMA-2 7B model (without fine-tuning) performs the poorest among all tested models, with only 48\% accuracy on the tested dataset. The huge accuracy gap between our fine-tuned model and the foundational LLaMA-2 7B indicates the effectiveness of our method when applied to the LLaMA-2 7B model. Other baseline models of approximately 7B parameter size, such as ChatGLM and BaiChuan, also demonstrate better performance than LLaMA-2 7B, but still fall short of our model, with accuracies ranging from 66\% to 70\%.

For the hard testing benchmark that involves complex networking analysis and logical reasoning, our model surpasses all the baseline models with an accuracy of 69\%, again demonstrating superior performance than other models tested. Other 7B models show a significant decrease in accuracy (compared to the easy testing benchmark) due to the increased problem difficulty, with accuracies generally falling between 55\% and 63\%. For complex problems, LLaMA-2 70B demonstrates the second-best performance among all models tested and shows a substantial lead over the third-best model (i.e., ChatGLM3-6B). This observation aligns well with previously reported observations \cite{kaplan2020scaling} about the relationship between LLM’s reasoning proficiency and model size, i.e., language models with larger sizes are expected to have strong reasoning abilities in general. Our model, despite having only one-tenth the parameters of LLaMA-2 70B, still marginally outperforms the larger model. This experimental observation serves as a compelling illustration of the enhanced reasoning capabilities facilitated by the RaC algorithm.

In the comprehensive benchmark, which contains a mix of easy and hard tasks, our method maintains its leading position with an accuracy of roughly 75\%. The LLaMA-2 70B model follows closely behind, achieving about 72\% accuracy. Other models in this testing benchmark show varying levels of performance, with most falling in the 65-70\% accuracy range. Our model’s superiority in effectively addressing these multifaceted questions suggests a potential for the model's application in solving real-world problems, which require both solid knowledge understanding and logical reasoning.

To conclude, we emphasize the effectiveness of our RaC fine-tuning framework by highlighting the following key experimental observations in the three testing benchmarks:
\begin{enumerate}
    \item Our model outperforms all baseline models tested. It not only outperforms the largest foundation model of the same family (i.e., the LLaMA-2 series) with only one-tenth of the parameter size, but it also outperforms all later proposed models with similar sizes.
    \item There is an obvious increase in accuracy between our fine-tuned model and the original foundational model (i.e., LLaMA-2 7B). The accuracy increases from 47\% to 80\%, 41\% to 68\%, and 45\% to 74\% on the easy benchmark, hard benchmark, and comprehensive dataset, respectively.
\end{enumerate}

\begin{figure}[htbp]
  \centering
  \includegraphics[width=0.46\textwidth]{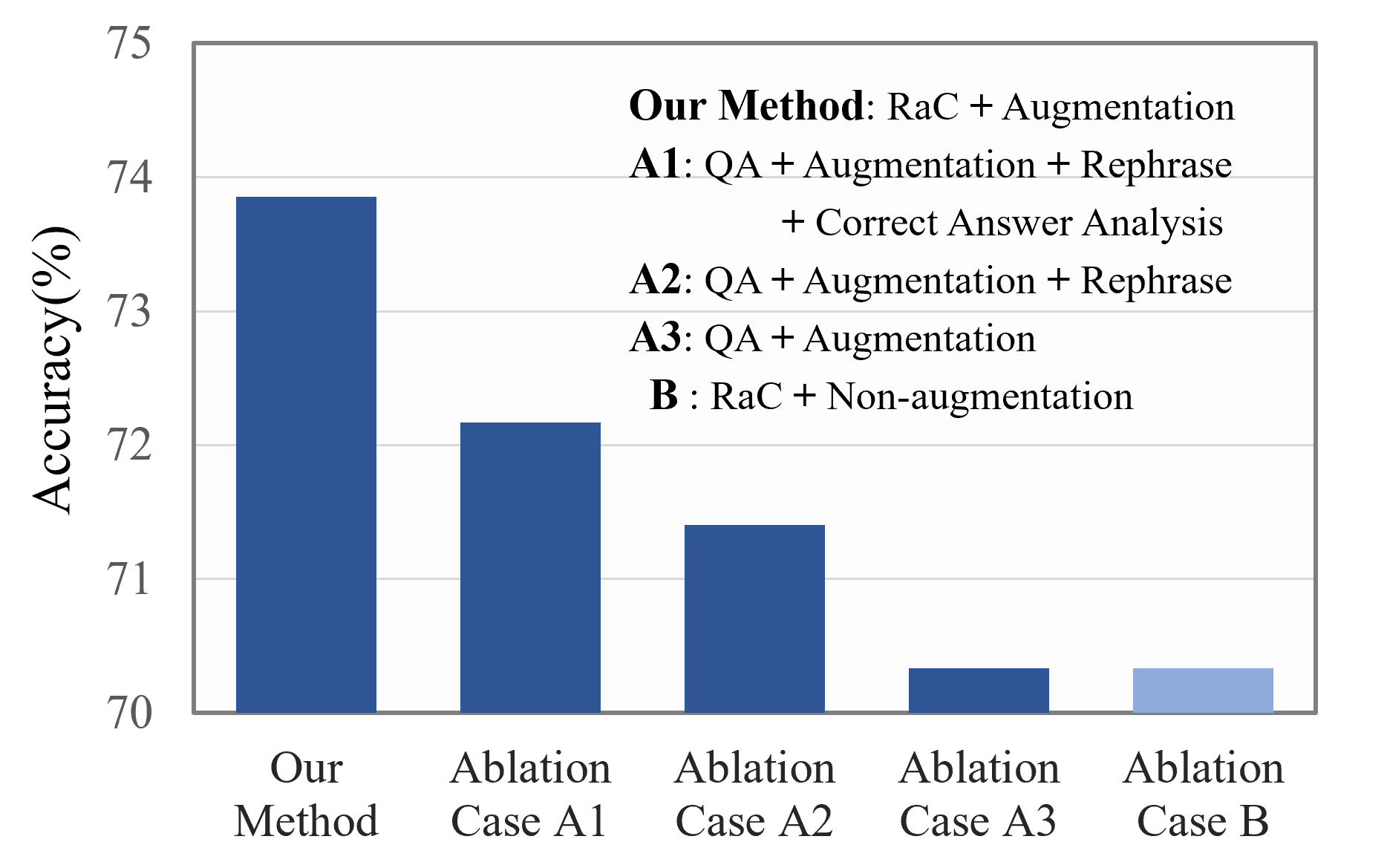}\\
  \captionsetup{font={footnotesize}}
  \caption{Accuracies of our model, ablation case A1 (our model - wrong answer explanation), A2 (A1 - correct answer explanation), A3 (A2 - question rephrasing) on the comprehensive dataset.}\label{fig6}
\end{figure}

To assess the efficacy of individual components within our methodology, we conducted a series of ablation studies. The Rephrase and Compare (RaC) framework comprises four key elements: (1) question-answer pairs, (2) problem rephrasing, (3) correct answer explanations, and (4) incorrect answer explanations. Our ablation experiments involve systematic removals of these components, followed by an evaluation of the resultant model's performance on a comprehensive testing benchmark. Specifically, we implemented three ablation cases: A1, which eliminated incorrect answer explanations; A2, which further removed correct answer explanations from A1; and A3, which excluded the problem rephrasing component from A2, leaving only question-answer pairs in the fine-tuning process. As illustrated in Fig. \ref{fig6}, the sequential elimination of these elements from the RaC framework led to a progressive decline in performance, underscoring the significance of each component within the framework.

Additionally, we examined an alternative ablation case (designated as case B) that omitted the data augmentation method delineated in Section \ref{sec-IV}. A comparative analysis between our proposed method and ablation case B revealed a substantial reduction in the model's answer accuracy. This empirical observation emphasizes the critical role of our data augmentation scheme in enhancing overall performance.

%% file: Sources/Section_VI.tex
This paper introduces RaC, an efficient fine-tuning framework for LLM. RaC addresses the limitations arising from over-reliance on prompting techniques and the lack of effective fine-tuning methods for lightweight models. By incorporating question reformulation and contrastive analysis during fine-tuning, RaC significantly enhances LLMs' comprehension and critical thinking abilities within networking contexts. Our experimental results demonstrate a 63.73\% accuracy improvement over the foundational model when tested on a comprehensive networking problem set. Furthermore, to obtain the dataset used in RaC fine-tuning, we develop a GPT-assisted data mining method for generating high-quality RaC QA pairs and introduce ChoiceBoost, a data augmentation technique that expands dataset size while reducing answer bias. 

For model evaluation, we introduce three new testing benchmarks of varying difficulty, which provide a standardized means of assessing network-oriented LLMs. Our open-source contributions, including the training dataset, three testing benchmarks, the fine-tuned model, and associated codes, facilitate the reproducibility of our work and encourage further advancements in this field.

This work marks a significant step towards more effective LLMs for networking applications, bridging the gap between general-purpose language models and domain-specific networking knowledge. Future work could explore the scalability of RaC to larger models and diverse networking tasks, investigate transfer learning between different networking domains, develop more sophisticated data augmentation techniques, and conduct user studies in real-world scenarios.

%% file: main.bbl
\begin{thebibliography}{10}
\providecommand{\url}[1]{#1}
\csname url@samestyle\endcsname
\providecommand{\newblock}{\relax}
\providecommand{\bibinfo}[2]{#2}
\providecommand{\BIBentrySTDinterwordspacing}{\spaceskip=0pt\relax}
\providecommand{\BIBentryALTinterwordstretchfactor}{4}
\providecommand{\BIBentryALTinterwordspacing}{\spaceskip=\fontdimen2\font plus
\BIBentryALTinterwordstretchfactor\fontdimen3\font minus \fontdimen4\font\relax}
\providecommand{\BIBforeignlanguage}[2]{{%
\expandafter\ifx\csname l@#1\endcsname\relax
\typeout{** WARNING: IEEEtran.bst: No hyphenation pattern has been}%
\typeout{** loaded for the language `#1'. Using the pattern for}%
\typeout{** the default language instead.}%
\else
\language=\csname l@#1\endcsname
\fi
#2}}
\providecommand{\BIBdecl}{\relax}
\BIBdecl

\bibitem{niu2024large}
Q.~Niu, J.~Liu, Z.~Bi, P.~Feng, B.~Peng, and K.~Chen, ``Large language models and cognitive science: A comprehensive review of similarities, differences, and challenges,'' \emph{arXiv preprint arXiv:2409.02387}, 2024.

\bibitem{he2024llm}
Z.~He, A.~Gottipati, L.~Qiu, F.~Y. Yan, X.~Luo, K.~Xu, and Y.~Yang, ``Llm-abr: Designing adaptive bitrate algorithms via large language models,'' \emph{arXiv preprint arXiv:2404.01617}, 2024.

\bibitem{mani2023enhancing}
S.~K. Mani, Y.~Zhou, K.~Hsieh, S.~Segarra, T.~Eberl, E.~Azulai, I.~Frizler, R.~Chandra, and S.~Kandula, ``Enhancing network management using code generated by large language models,'' in \emph{Proceedings of the 22nd ACM Workshop on Hot Topics in Networks}, 2023, pp. 196--204.

\bibitem{mondal2023llms}
R.~Mondal, A.~Tang, R.~Beckett, T.~Millstein, and G.~Varghese, ``What do llms need to synthesize correct router configurations?'' in \emph{Proceedings of the 22nd ACM Workshop on Hot Topics in Networks}, 2023, pp. 189--195.

\bibitem{lian2023configuration}
X.~Lian, Y.~Chen, R.~Cheng, J.~Huang, P.~Thakkar, and T.~Xu, ``Configuration validation with large language models,'' \emph{arXiv preprint arXiv:2310.09690}, 2023.

\bibitem{cui2023llmind}
H.~Cui, Y.~Du, Q.~Yang, Y.~Shao, and S.~C. Liew, ``Llmind: Orchestrating ai and iot with llms for complex task execution,'' \emph{arXiv preprint arXiv:2312.09007}, 2023.

\bibitem{wu2024netllm}
D.~Wu, X.~Wang, Y.~Qiao, Z.~Wang, J.~Jiang, S.~Cui, and F.~Wang, ``Netllm: Adapting large language models for networking,'' in \emph{Proceedings of the ACM SIGCOMM 2024 Conference}, 2024, pp. 661--678.

\bibitem{shao2024wirelessllm}
J.~Shao, J.~Tong, Q.~Wu, W.~Guo, Z.~Li, Z.~Lin, and J.~Zhang, ``Wirelessllm: Empowering large language models towards wireless intelligence,'' \emph{arXiv preprint arXiv:2405.17053}, 2024.

\bibitem{kotaru2023adapting}
M.~Kotaru, ``Adapting foundation models for operator data analytics,'' in \emph{Proceedings of the 22nd ACM Workshop on Hot Topics in Networks}, 2023, pp. 172--179.

\bibitem{zhou2023towards}
Y.~Zhou, N.~Yu, and Z.~Liu, ``Towards interactive research agents for internet incident investigation,'' in \emph{Proceedings of the 22nd ACM Workshop on Hot Topics in Networks}, 2023, pp. 33--40.

\bibitem{meng2024large}
R.~Meng, M.~Mirchev, M.~B{\"o}hme, and A.~Roychoudhury, ``Large language model guided protocol fuzzing,'' in \emph{Proceedings of the 31st Annual Network and Distributed System Security Symposium (NDSS)}, 2024.

\bibitem{zou2024promptintern}
J.~Zou, M.~Zhou, T.~Li, S.~Han, and D.~Zhang, ``Promptintern: Saving inference costs by internalizing recurrent prompt during large language model fine-tuning,'' \emph{arXiv preprint arXiv:2407.02211}, 2024.

\bibitem{nikbakht2024tspec}
R.~Nikbakht, M.~Benzaghta, and G.~Geraci, ``Tspec-llm: An open-source dataset for llm understanding of 3gpp specifications,'' \emph{arXiv preprint arXiv:2406.01768}, 2024.

\bibitem{maatouk2023teleqna}
A.~Maatouk, F.~Ayed, N.~Piovesan, A.~De~Domenico, M.~Debbah, and Z.-Q. Luo, ``Teleqna: A benchmark dataset to assess large language models telecommunications knowledge,'' \emph{arXiv preprint arXiv:2310.15051}, 2023.

\bibitem{kaplan2020scaling}
J.~Kaplan, S.~McCandlish, T.~Henighan, T.~B. Brown, B.~Chess, R.~Child, S.~Gray, A.~Radford, J.~Wu, and D.~Amodei, ``Scaling laws for neural language models,'' \emph{arXiv preprint arXiv:2001.08361}, 2020.

\end{thebibliography}
